\documentclass[10pt,twocolumn,letterpaper]{article}

\usepackage{iccv}
\usepackage{times}
\usepackage{epsfig}
\usepackage{graphicx}
\usepackage{amsmath}
\usepackage{amssymb}
\usepackage{subfigure}

\usepackage[breaklinks=true,bookmarks=false]{hyperref}

\iccvfinalcopy % *** Uncomment this line for the final submission

\def\iccvPaperID{****} % *** Enter the ICCV Paper ID here

\begin{document}

%%%%%%%%% TITLE
\title{Learning Multi-Scene Absolute Pose Regression with Transformers}

\author{Yoli Shavit
\and Ron Ferens\\
Bar-Ilan University, Israel
\and Yosi Keller
}

\maketitle
%\ificcvfinal\thispagestyle{empty}\fi

%%%%%%%%% ABSTRACT
\begin{abstract}
   Absolute camera pose regressors estimate the position and orientation of a
camera from the captured image alone. Typically, a convolutional backbone
with a multi-layer perceptron head is trained with images and pose labels to
embed a single reference scene at a time. Recently, this scheme was extended
for learning multiple scenes by replacing the MLP head with a set of fully
connected layers. In this work, we propose to learn multi-scene absolute
camera pose regression with Transformers, where encoders are used to
aggregate activation maps with self-attention and decoders transform latent
features and scenes encoding into candidate pose predictions. This mechanism
allows our model to focus on general features that are informative for
localization while embedding multiple scenes in parallel. We evaluate our
method on commonly benchmarked indoor and outdoor datasets and show that it
surpasses both multi-scene and state-of-the-art single-scene absolute pose
regressors. We make our code publicly available from \href{https://github.com/yolish/multi-scene-pose-transformer}%
{https://github.com/yolish/multi-scene-pose-transformer}.
\end{abstract}

%%%%%%%%% BODY TEXT
%%%%%%%%% BODY TEXT

\section{Introduction}
\label{sec:methods}
\begin{figure}[h]
\includegraphics[width=\linewidth]{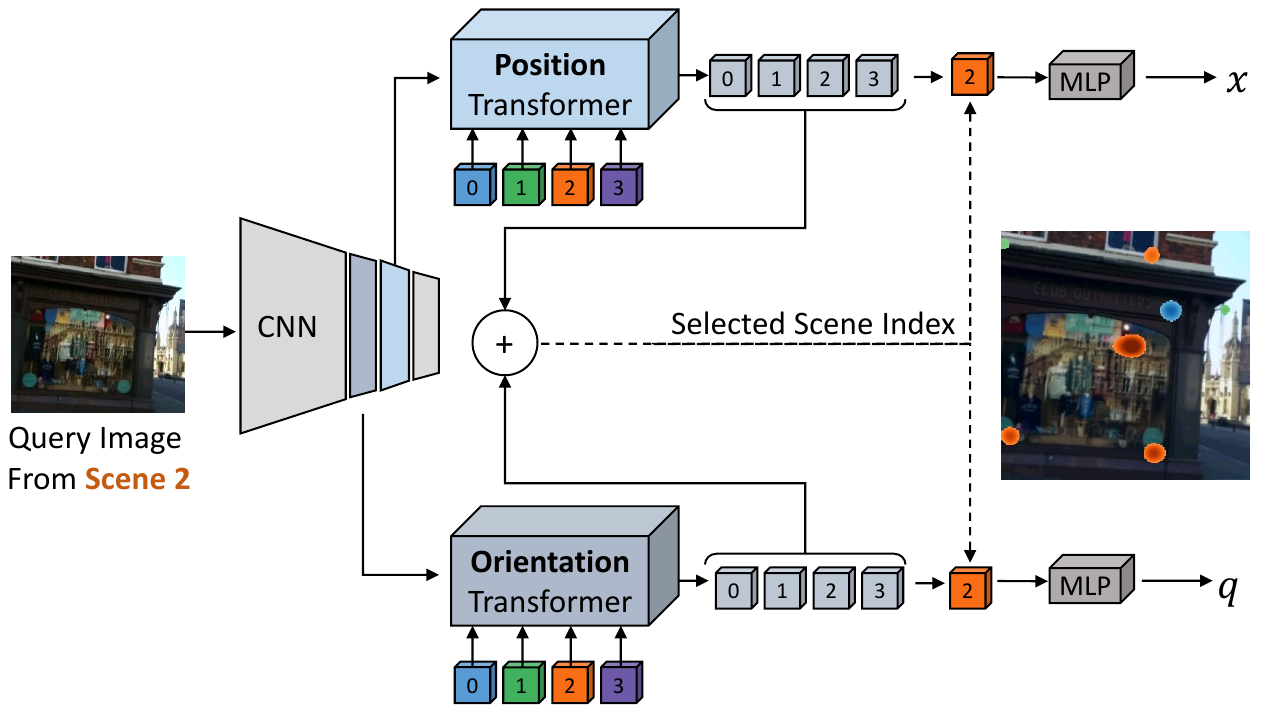}

\caption{Multi-scene absolute pose regression with Transformers. Two Transformers separately attend to position-  and orientation- informative features from a convolutional backbone. Scene-specific queries ($0-3$) are further encoded with aggregated activation maps into latent representations, from which a
single output is selected. The strongest response, shown as an overlaid color-coded heatmap of attention weights, is obtained with the output associated with the input image's scene. The selected outputs are used to regress the position $x$ and the orientation $q$.}
\label{fig:teaser}
\end{figure}
Localizing a camera using a query image is a key task in many computer vision
applications, such as indoor navigation, augmented reality and autonomous
driving, to name a few. Contemporary approaches for estimating the position
and orientation of a camera offer different trade-offs between accuracy,
runtime and memory. For example, hierarchical localization pipelines \cite%
{sattler2016efficient,taira2018inloc,sarlin2019coarse} achieve
state-of-the-art (SOTA) pose accuracy, but are relatively slow (response time
of hundreds of milliseconds) and require a large memory footprint and
client-server connectivity. These approaches employ image retrieval (IR) to
fetch images that are similar to the query image, followed by local features
extraction and matching. The extracted 2D-2D matches are mapped to 2D-3D
correspondences via depth or a 3D point cloud, and are then used to estimate
the camera pose with Perspective-n-Point (PnP) and RANSAC  \cite%
{fischler1981random}. Absolute pose regressors (APRs), on the other hand,
estimate the camera pose with a single forward pass, using only the query
image. They are an order of magnitude faster and can be deployed as a
standalone application on a thin client thanks to their small memory
footprint. Unfortunately, APRs are also an order of magnitude less accurate
compared to hierarchical localization pipelines and to other methods
utilizing 3D data at inference time \cite{sattler2019understanding}. Moreover,
most APRs are designed to embed a single scene at a time, implying that for
a dataset with $N$ scenes (for example, a hospital with many wards and
rooms), $N$ models are required to be trained, deployed and chosen from
during inference. In this work, we focus on improving the accuracy of APRs
while extending the current single-scene paradigm for learning multiple
scenes in parallel. \newline
The formulation of absolute camera pose regression was first suggested by
Kendall et al. \cite{kendall2015posenet}. Following the success of
convolutional neural networks (CNNs) in learning different computer vision
tasks, the authors suggested to adapt a GoogLeNet architecture to camera
pose regression by attaching a multi-layer perceptron (MLP) head to regress
the camera position and orientation. The proposed architecture, denoted
PoseNet, offered a novel, fast and lightweight solution for camera
localization. However, it also suffered from low accuracy and limited
generalization. Various absolute pose regression methods were suggested to
address these issues, proposing modifications to the backbone and MLP architecture \cite%
{melekhov2017image,naseer2017deep,walch2017image,wu2017delving,shavitferensirpnet,wang2020atloc,cai2019hybrid}%
, as well as different loss formulations and optimization strategies \cite%
{kendall2016modelling,kendall2017geometric, shavit2019introduction}. Despite
their variety, such APRs share two common traits: (1) employing a CNN
backbone to output a single global latent vector which is used for
regressing the pose (2) training a model per scene (scene-specific APRs).
\newline
Recently, Blanton et al. \cite{blanton2020extending} suggested a method for
extending single-scene absolute pose regression to a multi-scene paradigm.
Similarly to existing APRs, this method applies a CNN backbone for
generating a latent global descriptor of the image. However, instead of
using a single scene-specific MLP, it trains a set of Fully Connected (FC)
layers, with a layer per scene, which is indexed based on the predicted
scene identifier. While offering a new general framework for optimizing a
single model for multiple scenes, this method was unable to match the
accuracy of contemporary SOTA APRs.\newline
In this work, we propose a novel formulation of multi-scene absolute pose
regression, inspired by recent successful applications of Transformers to
computer vision tasks such as object detection \cite{DETR} and image
recognition \cite{16x16}. These works demonstrated the effectivity of
\textit{encoders} in focusing on latent features (in image patches or
activation maps) that are informative for particular tasks, through
self-attention aggregation. In addition, \textit{decoders} were shown to
successfully generate multiple independent predictions, corresponding to
queries, based on the input embedding \cite{DETR}. \ Similarly, we propose
to apply Transformers to multi-scene absolute pose regression, using \textit{%
encoders} to focus on pose-informative features and \textit{decoders} to
transform encoded scene identifiers to latent pose representations (Fig. \ref%
{fig:teaser}). As pose estimation involves two different tasks (position and
orientation estimation), related to different visual cues, we apply a shared
convolutional backbone at two different resolutions and use two different
Transformers, one per task. The decoders' outputs are used to classify the
scene and select the respective position and orientation embeddings, from
which the position and orientation vectors are regressed. \newline
We evaluate our approach on two commonly benchmarked datasets, consisting of
multiple outdoor and indoor localization challenges. We show that our method
not only provides a new SOTA accuracy for \textit{multi-scene APR}
localization, but importantly provides a new SOTA for \textit{single-scene}
\textit{APRs}. Moreover, we show that our approach achieves competitive
results even when trained across multiple datasets of significantly
different characteristics. We further conduct multiple ablations to evaluate
the sensitivity of our model to different design choices and analyze its
scalability in terms of runtime and memory. In summary, our main
contributions are as follows:

\begin{itemize}
\item We propose a novel formulation for multi-scene absolute pose
regression using Transformers.

\item We experimentally demonstrate that self-attention allows aggregation of
positional and rotational image cues.

\item Our approach is shown to achieve new SOTA accuracy for both
multi-scene and single-scene APRs across contemporary outdoor and indoor
localization benchmarks.
\end{itemize}

%------------------------------------------------------------------------

\section{Related Work}
\label{sec:related}
Camera pose estimation methods can be divided into several families, depending on the inputs at inference time and on their algorithmic characteristics.\\
\textbf{Image Retrieval} methods learn global image descriptors for
retrieving database images that depict the vicinity of the area captured by
the query image. They are commonly employed by pose estimation methods such
as structure-based hierarchical localization pipelines \cite%
{taira2018inloc,sarlin2019coarse,noh2017large,dusmanu2019d2} and relative
pose regression methods \cite%
{balntas2018relocnet,laskar2017camera,ding2019camnet}. IR can also be
applied for estimating the camera pose by taking the pose of the most similar
fetched image, or by interpolating the poses of several visually close
images. Such approaches require both storing and searching through large
databases with pose labels. Recently, Sattler et al. \cite%
{sattler2019understanding} proposed an IR-based baseline for camera pose
regression to illustrate the limitations of APRs, as no regressor was able
to consistently surpass it on multiple localization tasks.\\
\textbf{3D-based Localization} methods, also referred to as structure-based
methods \cite{sattler2016efficient,sattler2019understanding}, include
camera pose estimation techniques that utilize the correspondences between
2D image positions and 3D world coordinates for camera localization with
PnP and RANSAC. Hierarchical pose estimation pipelines \cite%
{taira2018inloc,sarlin2019coarse,noh2017large,dusmanu2019d2} are based on a
two-phase approach, utilizing global (IR) and local matching. Each query
image to be localized, is first encoded using a CNN trained for IR, and a
relatively small set of nearest neighbors is retrieved from a large-scale
image dataset. Tentative 2D-2D correspondences are estimated by matching
local image features, and are then mapped into 2D-3D matches. The resulting
matches are passed to PnP-RANSAC for estimating the camera pose. Such
pipelines were shown to achieve SOTA accuracy on large-scale benchmarks with
challenging conditions \cite{taira2018inloc,sarlin2019coarse}. However, these approaches are slower than other localization methods by an order or two
orders of magnitude and are typically deployed with a client-server
architecture due to the large storage space needed. A different body of
works directly regresses the 3D coordinates from 2D positions in the image.
Brachmann and Rother derived the DSAC \cite{DSAC} and the follow-up DSAC++ \cite%
{DSAC++} schemes, where a CNN architecture is trained to estimate the 3D
locations of the pixels in the query image, in order to establish 2D-3D
correspondences for estimating the camera pose with PnP-RANSAC. These
methods only require the query image at inference time, and achieve SOTA
accuracy, which is competitive with hierarchical localization pipelines.
However, similar to APRs, a model needs to be trained per scene. In
addition, these method are challenging to implement, require a long time to
converge and are slower (100ms) by an order of magnitude compared to
absolute pose regression approaches (10ms) at inference time \cite%
{blanton2020extending}. They also suffer from a non-deterministic behavior due to the
inherent randomness of RANSAC. \newline
\textbf{Relative Pose Regression} methods combine camera pose regression
with an IR scheme. The absolute camera pose is computed by first estimating
the \textit{relative} motion (translation and rotation) between the query
image and a set of reference images, for which the ground truth pose is
known. An IR scheme is applied to retrieve a set of nearest neighbors
images, and a relative motion regression is separately computed between the
query image and each of the retrieved images, followed by pose
interpolation. The learning thus focuses on regressing the relative pose
given a pair of images \cite%
{balntas2018relocnet,laskar2017camera,ding2019camnet}. These relative pose
regressors (RPRs) were shown to generalize better than APRs and to improve the accuracy on
small scale indoor benchmarks \cite{ding2019camnet}. However, similar to
other IR-based schemes, such approaches require a retrieval phase and a
pose-labelled database during inference time. When the images are
sequentially acquired over time, combining relative and absolute regression
was shown to significantly improve the pose accuracy \cite%
{valada2018deep,radwan2018vlocnet++}.\newline
\textbf{Absolute Pose Regression} was first proposed by \cite%
{kendall2015posenet} to directly regress the position and orientation of the
camera, given the input image, by attaching an MLP head to a GoogLeNet
backbone. The resulting architecture, named PoseNet, was far less accurate
than 3D-based methods, but enabled pose estimation using a single forward
pass, offering a much lighter and faster localization alternative. In order
to improve the localization accuracy, contemporary APRs studied different
CNN backbones \cite{melekhov2017image,naseer2017deep,wu2017delving,
shavitferensirpnet} and MLP heads \cite{wu2017delving,naseer2017deep}.
Overfitting was addressed by averaging the predictions of multiple models
with randomly dropped activations \cite{kendall2016modelling}, or by
reducing the dimensionality of the global image encoding using
Long-Short-Term-Memory (LSTM) layers \cite{walch2017image}. Other works
focused on the loss formulation for absolute pose regression in order to
enable adaptive weighing of the position and orientation associated errors.
Kendall et al. \cite{kendall2017geometric} suggested to optimize the
parameters balancing both losses for improving accuracy and avoiding manual fine-tuning. This formulation was
adopted by many pose regressors. Alternative orientation representations
were also proposed to improve the pose loss \cite%
{wu2017delving,brahmbhatt2018geometry}. The use of additional sensors, such
as inertial sensors, was also suggested to improve the localization accuracy
\cite{brahmbhatt2018geometry}. More recently, Wang et al. \cite%
{wang2020atloc} proposed to use attention for guiding the regression process
by applying self-attention on the output of the CNN backbone. The new
attention-based representation was used to regress the pose with an MLP
head. While many modifications were proposed to the architecture and loss
originally formulated by Kendall et al., the main paradigm remained the
same: (1) employing a CNN backbone to output a global latent vector which is
used for absolute pose regression (2) training a separate model per scene.
\newline
\textbf{Multi-Scene Absolute Pose Regression} methods aim to extend the
absolute pose regression paradigm for learning a \textit{single} model on
\textit{multiple} scenes. Blanton et al. proposed the Multi-Scene PoseNet
(MSPN), a novel multi-scene absolute pose regression approach \cite%
{blanton2020extending}, where the network first classifies the particular scene
related to the input image, and then uses it to index a set of
scene-specific weights for regressing the pose. An activation map from a CNN
backbone, which is shared across scenes, is used both for scene
classification and regressing the pose. A fully connected layer with SoftMax
predicts the scene and is trained via Binary Cross Entropy. A set of FC
layers, one per scene, is trained for absolute camera pose regression with a
set of scene-specific parameterized losses. The notion of multi-scene camera
pose estimation was also applied to 3D-based methods which regresses the 3D
coordinates from image pixels. However, the suggested framework still
involved training multiple models (one per scene) and then selecting the
most appropriate model using a mixture-of-experts strategy \cite%
{brachmann2019expert}. \newline
In this work, we focus on learning a \textit{single} unified deep learning
model for absolute pose regression across multiple scenes. Our method is
thus closely related to single- and multi-scene absolute pose regression and
we compare it to leading architectures (APRs) in this field. We include additional background on Attention and Transformers in the supplementary material.
\section{Multi-Scene Absolute Camera Pose Regression with Transformers}
\label{sec:methods}
\begin{figure*}[th]
\begin{center}
\includegraphics[width=0.9\textwidth]{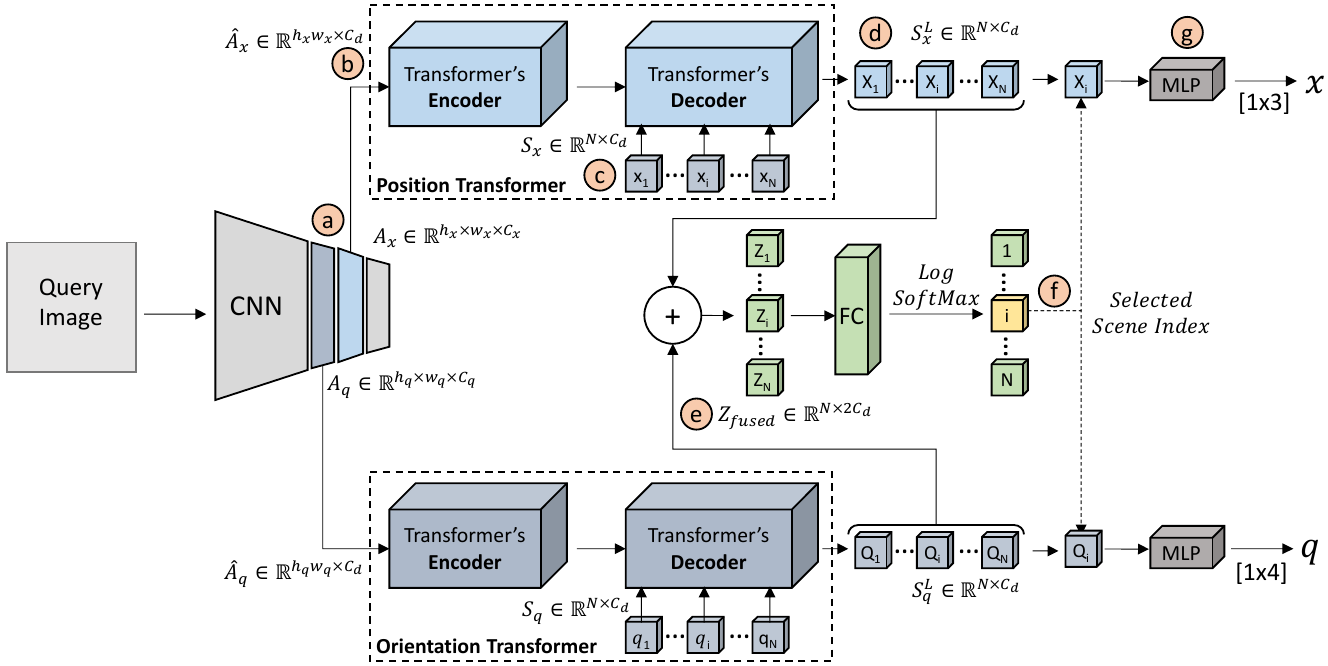}
\end{center}
\caption{The architecture of our proposed model.}
\label{fig:architecture}
\end{figure*}
A single/multi-scene APR localizes the capturing camera with a forward pass
on the input image. The camera pose $\mathbf{p}$ is typically represented by
the tuple $<\mathbf{x},\mathbf{q}>$ where $\mathbf{x}\in \mathbb{R}^{3}$ is
the position of the camera in the world coordinates and $\mathbf{q}\in
\mathbb{R}^{4}$ is the quaternion encoding of its 3D\ orientation. Following
the success of recent visual Transformers \cite{DETR,16x16}, we employ
separate positional and orientational \textit{Transformer Encoders} for
adaptive aggregation of (flattened) intermediate activation maps computed by a convolutional backbone. In particular, as depicted in Fig. \ref%
{fig:att-enocder}, the positional and orientational encoders emphasize
different image cues:\ corner and blob-like image cues are
position-informative, in contrast to the elongated edges emphasized by
the orientational encoder. In order to attend to $N$ scenes we also apply
separate positional and orientational \textit{Transformer Decoders}, that
are queried by $\left\{ \mathbf{x}_{i}\right\} _{1}^{N}$ and $\left\{
\mathbf{q}_{i}\right\} _{1}^{N},$ for the position and orientation
embeddings per scene, respectively. The corresponding output sequences, $%
\left\{ \mathbf{X}_{i}\right\} _{1}^{N}$ and $\left\{ \mathbf{Q}_{i}\right\}
_{1}^{N}$, respectively, encode the localization parameters \emph{per scene}%
. This architecture is inspired by the DETR approach \cite{DETR}, where a
single activation map is queried by multiple queries, each related to a
different task. Together, the encoder-decoder Transformer architecture
allows to attend to localization-informative image content while learning
multiple scenes at once. In order to regress the pose, the scene is first
classified by concatenating $\left\{ \mathbf{X}_{i}\right\} _{1}^{N}$ and $%
\left\{ \mathbf{Q}_{i}\right\} _{1}^{N}$, and the embeddings of the detected
scene $\left\{ \mathbf{X}_{i},\mathbf{Q}_{i}\right\} $ are regressed by the
MLP heads.

\subsection{Network Architecture}

\label{subsec:model_architecture}

The architecture of our model is shown in Fig. \ref{fig:architecture}. Given
an image $\mathbf{I}\in \mathbb{R}^{H\times W\times C}$, we sample a
convolutional backbone at two different resolutions and take an activation
map per regression task: $A_{\mathbf{x}}$ and $A_{\mathbf{q}}$, for position
and orientation regression, respectively (Fig. \ref{fig:architecture}a). In
order to transform activation maps into Transformer-compatible inputs, we
follow the same sequence preparation procedure as in \cite{DETR}. An
activation map $\mathbf{A}\in \mathbb{R}^{H_{a}\times W_{a}\times C_{a}}$ is
first converted to a sequential representation $\widehat{\mathbf{A}}\in
\mathbb{R}^{{H_{a}}\cdot W_{a}\times C_{d}}$ using a $1\times 1$ convolution
(projecting to dimension $C_{d}$) followed by flattening (Figure \ref%
{fig:architecture}b). Each position in the activation map is further
assigned with a learned encoding to preserve the spatial information of each
location. In order to reduce the number of parameters, two one-dimensional
encodings are separately learned for the $X$,$Y$ axes. Specifically, for an
activation map $\mathbf{A}$ we define the sets of positional embedding
vectors $\mathbf{E}_{u}\in \mathbb{R}^{\left( W_{a}\right) \times C_{d}/2}$
and $\mathbf{E}_{v}\in \mathbb{R}^{\left( H_{a}\right) \times C_{d}/2}$,
such that a spatial position $\left( i,j\right) ,$ $i\in 1..H_{a}$, $j\in
1..W_{a}$, is encoded by the concatenating of the two corresponding
embedding vectors:%
\begin{equation}
\mathbf{E}_{pos}^{i,j}=%
\begin{bmatrix}
\mathbf{E}_{u}^{j} \\
\mathbf{E}_{v}^{i}%
\end{bmatrix}%
\in \mathbb{R}^{C_{d}}.
\end{equation}%
The processed sequence, serving as input to the Transformer is thus given
by:
\begin{equation}
\mathbf{Z}_{\widehat{A}}^{0}=\widehat{\mathbf{A}}+\mathbf{E_{A}}\in \mathbb{R%
}^{{H_{a}}\cdot W_{a}\times C_{d}}
\end{equation}%
where $\mathbf{E_{A}}$ is the positional encoding of $A$. This processing is
applied separately for each of the two activation maps (for the position and
orientation Transformers, respectively).\newline
We use the Transformer architecture described in \cite{DETR}, using standard
Encoder and Decoder architectures modified to add the position encodings at
each attention layer. A Transformer Encoder is composed of $L$ identical
layers each consisting of multi-head attention (MHA) and multi-layer
perceptron (MLP) modules. Each layer $l$, $l=1..L$, performs the following
computation by applying a LayerNorm (LN) \cite{ba2016layer} before each
module and adding back the input with residual connections:
\begin{equation}
\mathbf{Z}_{\widehat{A}}^{l^{\prime }}=MHA(LN(\mathbf{Z}_{\widehat{A}%
}^{l-1}))+\mathbf{Z_{\widehat{A}}^{l-1}}\in \mathbb{R}^{{H_{a}}\cdot
W_{a}\times C_{d}}  \label{eq:encoder-mha}
\end{equation}%
\begin{equation}
\mathbf{Z}_{\widehat{A}}^{l}=MLP(LN(\mathbf{Z}_{\widehat{A}}^{l^{\prime }}))+%
\mathbf{Z_{\widehat{A}}^{l^{\prime }}}\in \mathbb{R}^{{H_{a}}\cdot
W_{a}\times C_{d}}  \label{eq:encoder-mlp}
\end{equation}%
At the final layer, $L$ the output is passed through an additional
normalization:
\begin{equation}
\mathbf{Z}_{\widehat{A}}^{L}=LN(\mathbf{Z}_{\widehat{A}}^{L})
\end{equation}

Given a dataset with $N$ scenes, the Transformer Decoders first applies self-attention as in Eq.~%
\ref{eq:encoder-mha} to the two learnt query sequences,
$\left\{ \mathbf{x}_{i}\right\} _{1}^{N}$ and $\left\{ \mathbf{q}%
_{i}\right\} _{1}^{N}$ (Fig. \ref{fig:architecture}c), for the position and orientation decoders,
respectively. Eqs.~%
\ref{eq:encoder-mha}-\ref{eq:encoder-mlp} are then applied again, but this time computing encoder-decoder attention instead of self-attention. As oppose to earlier auto-regressive decoders \cite%
{AttentionIsAllYouNeed}, this architecture outputs predictions in parallel
for all positions. We refer the reader to \cite{AttentionIsAllYouNeed,DETR}
for detailed definitions of the MHA operation and parallel decoding. The Transformers' Decoders
output the sequences $\left\{ \mathbf{X}_{i}\right\} _{1}^{N}$ and $\left\{
\mathbf{Q}_{i}\right\} _{1}^{N}$, with a latent embedding for each scene
(Fig. \ref{fig:architecture}d). However, given a query image, only one
position corresponds to the scene from which the image was taken. In order
to select the appropriate scene, we append the outputs of the two
transformers (Fig. \ref{fig:architecture}e) as $\left\{ \mathbf{Z}%
_{i}\right\} _{1}^{N}$ such that

\begin{equation}
\mathbf{Z}_{i}=%
\begin{bmatrix}
\mathbf{X}_{i} \\
\mathbf{Y}_{i}%
\end{bmatrix}%
\in \mathbb{R}^{2C_{d}},
\end{equation}%
and pass them through a fully connected layer followed by Log SoftMax. The
vectors at the position with the maximal probability are then chosen (Fig. %
\ref{fig:architecture}f). The selected Transformers outputs $\left\{ \mathbf{%
X}_{i},\mathbf{Q}_{i}\right\} $ are passed to a respective MLP head with one
hidden layer and gelu non-linearity to regress the target vectors, $\mathbf{x%
}$ (Fig. \ref{fig:architecture}g) or $\mathbf{q}$.

\subsection{Multi-Scene Camera Pose Loss}

We train our model to minimize both the position loss $L_{\mathbf{x}}$ and
the orientation loss $L_{\mathbf{q}}$, with respect to a ground truth pose $%
\mathbf{p}_{0}=<\mathbf{x}_{0},\mathbf{q}_{0}>$, given by:
\begin{equation}
L_{\mathbf{x}}=||\mathbf{x}_{0}-\mathbf{x}||_{2}  \label{equ:position loss}
\end{equation}%
\begin{equation}
L_{\mathbf{q}}=||\mathbf{q_{0}}-\frac{\mathbf{q}}{||\mathbf{q}||}||_{2}
\label{equ:orientation loss}
\end{equation}%
where $q$ is normalized to a unit norm quaternion in order ensure it is a
valid orientation encoding. We combine the two losses using the camera pose
loss formulation suggested by Kendall et al. \cite{kendall2017geometric}:
\begin{equation}
L_{\mathbf{p}}=L_{\mathbf{x}}\exp (-s_{\mathbf{x}})+s_{\mathbf{x}}+L_{%
\mathbf{q}}\exp (-s_{\mathbf{q}})+s_{\mathbf{q}}
\label{equ:learnable pose loss}
\end{equation}%
where $s_{\mathbf{x}}$ and $s_{\mathbf{q}}$ are learned parameters
controlling the balance between the two losses. Since our model is also
required to classify the scene from which the image was taken, we further
add a Negative Log Likelihood (NLL) loss term, computed with respect to the
ground truth scene index $s_{0}$. Given an estimated pose $p$ and the log
probability distribution $s$ of the predicted scene, our overall loss is
given by:
\begin{equation}
L_{\mathbf{multi-scene}}=L_{\mathbf{p}}+NLL(\mathbf{s},\mathbf{s_{0}})
\label{equ:multi-scene-loss}
\end{equation}

\subsection{Implementation Details}

Our model is implemented in PyTorch \cite{paszke2019pytorch}. We use a
pre-trained EfficientNet-B0 \cite{pmlr-v97-tan19a}, available through an
open source implementation \cite{efficientnet-pytorch}, and take $A_{\mathbf{%
x}}$ and $A_{\mathbf{q}}$ at two different resolutions: $A_{\mathbf{x}}\in
\mathbb{R}^{14\times 14\times 112}$ and $A_{\mathbf{q}}\in \mathbb{R}%
^{28\times 28\times 40}$. We set $C_{d}=256$ for the dimension of inputs of
the Transformer components. All encoders and decoders consist of six layers
with gelu non-linearity and with a dropout of $p=0.1$. Each
(encoder/decoder) layer uses four heads MHA and a two-layers MLP with a
hidden dimension $C_{h}=C_{d}$.\newline
The two MLP heads, regressing the position and orientation vectors,
respectively, expand the decoder dimension to $1024$ with a single hidden
layer. Our code is publicly available from \href{https://github.com/yolish/multi-scene-pose-transformer}%
{https://github.com/yolish/multi-scene-pose-transformer}, providing the model implementation along with a training and
evaluation framework.

\begin{table*}[th]
\caption{Comparative analysis of MSPN and our method on the Cambridge
Landmarks dataset (outdoor localization). We report the median
position/orientation error in meters/degrees for each method. Bold
highlighting indicates better performance.}
\label{tb:cambridge_multi_compare}\centering
\begin{tabular}{cccccc}
\hline
\textbf{Method} & \textbf{K. College} & \textbf{Old Hospital} & \textbf{Shop
Facade} & \textbf{St. Mary} &  \\ \hline
\multicolumn{1}{l}{MSPN \cite{blanton2020extending}} & 1.73/3.65 & 2.55/4.05
& 2.92/7.49 & 2.67/6.18 &  \\
\multicolumn{1}{l}{MS-Transformer (ours)} & \textbf{0.83}/\textbf{1.47} &
\textbf{1.81}/\textbf{2.39} & \textbf{0.86}/\textbf{3.07} & \textbf{1.62}/%
\textbf{3.99} &  \\ \hline
\end{tabular}%
\end{table*}

\begin{table*}[ht]
\caption{Comparative analysis of our method and MSPN on the 7Scenes dataset
(indoor localization). We report the median position/orientation error in
meters/degrees for each method. Bold highlighting indicates better
performance.}
\label{tb:7scenes_multi_compare}\centering
\begin{tabular}{cccccccc}
\hline
\textbf{Method} & \textbf{Chess} & \textbf{Fire} & \textbf{Heads} & \textbf{%
Office} & \textbf{Pumpkin} & \textbf{Kitchen} & \textbf{Stairs} \\ \hline
\multicolumn{1}{l}{MSPN\cite{blanton2020extending}} & \textbf{0.09}/4.76 &
0.29/10.5 & 0.16/13.1 & \textbf{0.16}/6.8 & 0.19/5.5 & \textbf{0.21}/6.61 &
0.31/11.63 \\
\multicolumn{1}{l}{MS-Transformer (ours)} & 0.11/\textbf{4.66} & \textbf{0.24%
}/\textbf{9.6} & \textbf{0.14}/\textbf{12.19} & 0.17/\textbf{5.66} & \textbf{%
0.18}/\textbf{4.44} & \textbf{0.17}/\textbf{5.94} & \textbf{0.26}/\textbf{%
8.45} \\ \hline
\end{tabular}%
\end{table*}
\begin{table}[ht]
\caption{Localization results for the Cambridge Landmarks dataset. We report
the average of median position/orientation errors in meters/degrees and the
respective rankings. Best results are highlighted in bold.}
\label{tb:cambridge_rank}\centering
\begin{tabular}{ccc}
\hline
\textbf{Method} & \textbf{Average} \textbf{[m/deg]} & \textbf{Ranks} \\
\hline
\multicolumn{3}{l}{Single-scene APRs} \\ \hline
\multicolumn{1}{l}{PoseNet \cite{kendall2015posenet}} & 2.09/6.84 & 10/11 \\
\multicolumn{1}{l}{BayesianPN \cite{kendall2016modelling}} & 1.92/6.28 & 8/10
\\
\multicolumn{1}{l}{LSTM-PN \cite{walch2017image}} & 1.30/5.52 & 2/9 \\
\multicolumn{1}{l}{SVS-Pose \cite{naseer2017deep}} & 1.33/5.17 & 3/7 \\
\multicolumn{1}{l}{GPoseNet \cite{cai2019hybrid}} & 2.08/4.59 & 6/3 \\
\multicolumn{1}{l}{PoseNet-Learnable \cite{kendall2017geometric}} & 1.43/2.85
& 5/2 \\
\multicolumn{1}{l}{GeoPoseNet \cite{kendall2017geometric}} & 1.63/2.86 & 6/3
\\
\multicolumn{1}{l}{MapNet \cite{brahmbhatt2018geometry}} & 1.63/3.64 & 6/5
\\
\multicolumn{1}{l}{IRPNet \cite{shavitferensirpnet}} & 1.42/3.45 & 4/4 \\
\hline
\multicolumn{3}{l}{Multi-scene APRs} \\ \hline
\multicolumn{1}{l}{MSPN \cite{blanton2020extending}} & 2.47/5.34 & 11/8 \\
\multicolumn{1}{l}{\textbf{MS-Transformer (Ours)}} & \textbf{1.28}/\textbf{%
2.73} & \textbf{1}/\textbf{1} \\ \hline
\end{tabular}%
\end{table}

\begin{table}[th]
\caption{Localization results for the 7Scenes dataset. We report the average
of median position/orientation errors in meters/degrees and the respective
rankings. Best results are highlighted in bold. }
\label{tb:7scenes_rank}\centering
\begin{tabular}{ccc}
\hline
\textbf{Method} & \textbf{Average} \textbf{[m/deg]} & \textbf{Ranks} \\
\hline
\multicolumn{3}{l}{Single-scene APRs} \\ \hline
\multicolumn{1}{l}{PoseNet \cite{kendall2015posenet}} & 0.44/10.4 & $10/11$
\\
\multicolumn{1}{l}{BayesianPN \cite{kendall2016modelling}} & 0.47/9.81 & $%
11/8$ \\
\multicolumn{1}{l}{LSTM-PN \cite{walch2017image}} & 0.31/9.86 & $8/9$ \\
\multicolumn{1}{l}{GPoseNet \cite{cai2019hybrid}} & 0.31/9.95 & $8/8$ \\
\multicolumn{1}{l}{PoseNet-Learnable \cite{kendall2017geometric}} & 0.24/7.87
& $7/4$ \\
\multicolumn{1}{l}{GeoPoseNet \cite{kendall2017geometric}} & 0.23/8.12 & $%
5/5 $ \\
\multicolumn{1}{l}{MapNet \cite{brahmbhatt2018geometry}} & 0.21/7.78 & $4/3$
\\
\multicolumn{1}{l}{IRPNet \cite{shavitferensirpnet}} & 0.23/8.49 & $5/7$ \\
\multicolumn{1}{l}{AttLoc \cite{wang2020atloc}} & 0.20/7.56 & 2/2 \\ \hline
\multicolumn{3}{l}{Multi-scene APRs} \\ \hline
\multicolumn{1}{l}{MSPN \cite{blanton2020extending}} & 0.20/8.41 & 2/6 \\
\multicolumn{1}{l}{\textbf{MS-Transformer (Ours)}} & \textbf{0.18}/ \textbf{%
7.28} & \textbf{1}/\textbf{1} \\ \hline
\end{tabular}%
\end{table}
\begin{table}[th]
\caption{Localization results with single-scene, multi-scene and
multi-dataset learning. We report the average of median position/orientation
errors in meters/degrees for the CambridgeLandmarks and 7Scenes datasets.}
\label{tb:multi_dataset}\centering
\begin{tabular}{lll}
\hline
\textbf{APR Method} & \textbf{CambridgeLand.} & \textbf{7Scenes} \\
\multicolumn{1}{c}{} & \multicolumn{1}{c}{\textbf{[m/deg]}} & \textbf{[m/deg]%
} \\ \hline
Single-scene \cite{kendall2017geometric} & \multicolumn{1}{c}{1.43/2.85} &
0.24/7.87 \\
Multi-scene (Ours) & \multicolumn{1}{c}{1.28/2.73} & 0.18/7.28 \\
Multi-dataset (Ours) & \multicolumn{1}{c}{1.50/ 2.57} & 0.22/6.78 \\ \hline
\end{tabular}%
\end{table}

\section{Experimental Results}

\subsection{Experimental Setup}

\textbf{Datasets.} We evaluate our approach using the Cambridge Landmarks
\cite{kendall2015posenet} and the 7Scenes \cite{glocker2013real} datasets,
which are commonly used for evaluating pose regression methods. The
Cambridge Landmarks dataset consists of six medium-sized scenes ($\sim
900-5500m^{2}$) set in an urban environment. For our comparative analysis,
we consider four scenes that are typically benchmarked by APRs. The 7Scenes
dataset includes seven small-scale scenes ($\sim 1-10m^{2}$) set in an
office indoor environment.\newline
\textbf{Training Details.} We optimize our model to minimize the loss in Eq. %
\ref{equ:multi-scene-loss} using Adam, with $\beta _{1}=0.9$, $\beta
_{2}=0.999$ and $\epsilon =10^{-10}$. The loss parameters (Eq. \ref%
{equ:learnable pose loss}) are initialized as in \cite{valada2018deep}.
Throughout all experiments we use a batch size of $8$ and an initial
learning rate of $\lambda =10^{-4}$. At train time, the decoder output is
selected using the ground truth scene index, and the estimated scene index
is used only for evaluating the NLL loss. During inference, the scene index
is not known and we rely on the prediction of our model, taking the index
with the highest (log) probability. Note that instead of training a model
per scene (typically with scene-specific training hyper-parameters) as in
single APRs, here we train a single model for \emph{all} the scenes
together. Additional details of augmentation and training are provided in
the supplementary material. All experiments reported in this paper were
performed on an 8Gb NVIDIA GeForce GTX 1080 GPU.

\subsection{Comparative Analysis of APRs}

\label{sec:comparative_analysis} Our method aims to offer a multi-scene
absolute pose regression paradigm which also improves the accuracy
obtained by current APRs. Hence, we compare our approach both to MSPN, which
is, to the best of our knowledge, the only other multi-scene APR, as well as
to leading single-scene APRs. We do not include the localization schemes detailed in
Section \ref{sec:related}, which are an order of magnitude slower
(3D-based scene coordinate regression \cite{DSAC,DSAC++}), or that utilize
additional data at inference time (localization pipelines \cite%
{taira2018inloc,sarlin2019coarse,noh2017large,dusmanu2019d2} and relative
pose regression \cite{balntas2018relocnet,laskar2017camera,ding2019camnet}).
Tables \ref{tb:cambridge_multi_compare} and \ref{tb:7scenes_multi_compare}
show the results obtained with our method (MS-Transformer) and with MSPN on
the CambridgeLandmarks and the 7Scenes datasets, respectively. Since MSPN
was trained on different scene combinations from the CambridgeLandmarks
dataset, we take the best performing model reported by the authors on this
dataset \cite{blanton2020extending}. Our method consistently outperforms
MSPN across outdoor and indoor scenes, reducing both position and
orientation errors.\newline
We further compare our results to contemporary absolute pose regression
solutions. Tables \ref{tb:cambridge_rank}-\ref{tb:7scenes_rank} show the
performance obtained by different APRs, MSPN and our method on the
CambridgeLandmarks and the 7Scenes datasets, respectively. We report the average
of median position and orientation errors across all scenes in each dataset
and the respective ranking (where top-1 corresponds to the smallest error).
Our method ranks first on both indoor and outdoor localization, achieving
the smallest position and orientation errors. Interestingly, the two best
performing APRs on the 7Scenes dataset (AttLoc and our method) both use the
attention mechanism for pose regression.\newline
We can further extend the notion of multi-scene learning to multi-dataset
learning, where a single model is trained on completely separate datasets,
potentially displaying different challenges and attributes. In order to
evaluate the effect of such an extension, we train our model on both the
7Scenes and Cambridge Landmarks datasets together. Table \ref%
{tb:multi_dataset} shows the average pose error per dataset for a \textit{%
state-of-the-art} single-scene APR \cite{kendall2017geometric} and our
model, trained in multi-scene and in multi-dataset modes. Although some
degradation is observed when training on both datasets together, our model
still maintains competitive performance and outperforms the single-scene
model. This is despite the two datasets depicting significantly different
environments and challenges (mid-scale outdoor versus small-scale indoor).
We also evaluate the ability of our model to correctly classify the scene of
the input query image. Our model achieves an average accuracy of $98.9\%$
(across scenes) allowing for a reliable selection of the decoder output,
which is essential for regressing the pose. Additional analysis of our model
scalability (runtime and memory) is provided in the supplementary material.
\subsection{Attention Maps Visualization and Interpretation}\label{sec:att-vis}
\begin{figure}[tbh]
\begin{center}
\subfigure[Input (3
scenes)]{\includegraphics[scale=0.25]{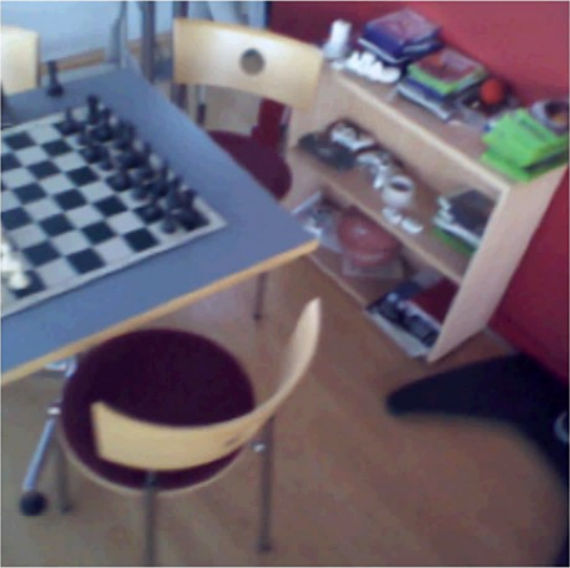}}
\subfigure[Position]{%
\includegraphics[scale=0.25]{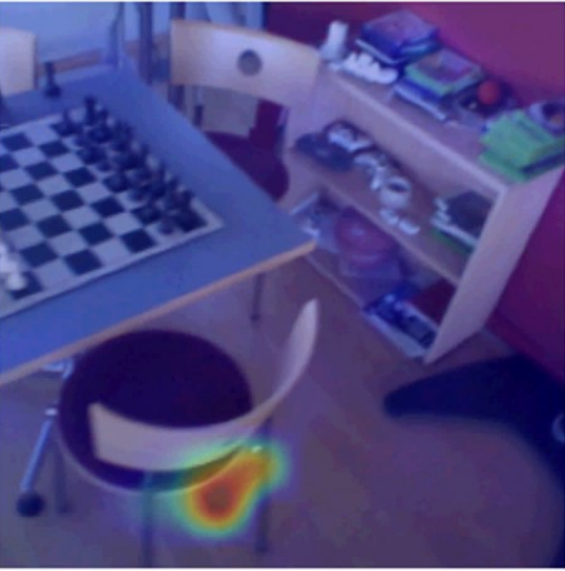}} %
\subfigure[Orientation]{%
\includegraphics[scale=0.25]{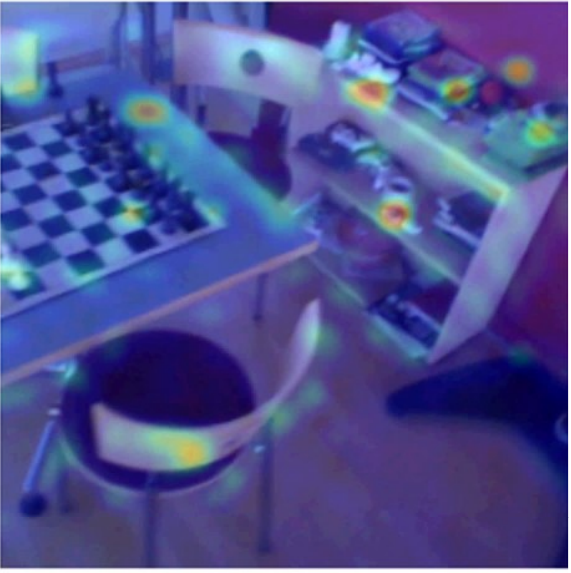}}
\subfigure[Input (7
scenes)]{\includegraphics[scale=0.25]{assets/chess_encoder_act_maps_comp1.pdf}}
\subfigure[Position]{%
\includegraphics[scale=0.25]{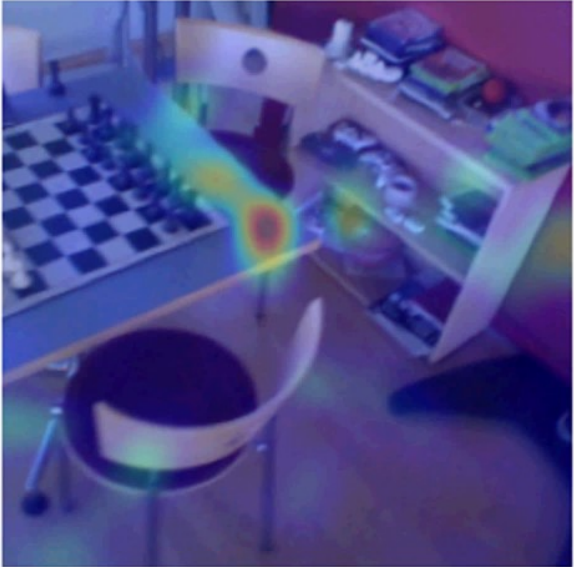}} %
\subfigure[Orientation]{%
\includegraphics[scale=0.25]{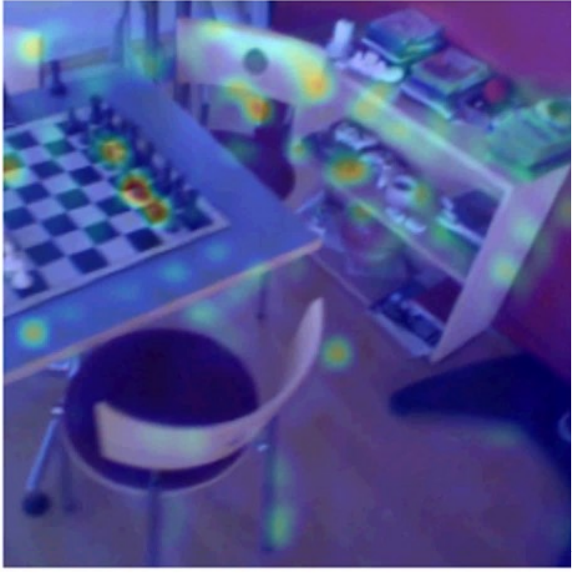}}
\end{center}
\caption{Transformer Encoder attention visualizations for a varying number
of training scenes (three and seven). As we train our scheme using more
scenes (second row), the positional attention is able to better localize
corner-like image cues (e compared to b). The orientational attention is able
to better localize elongated edges (f compared to c). }
\label{fig:att-enocder}
\end{figure}

\begin{figure}[tbh]
\begin{center}
\subfigure[Kings
C.]{\includegraphics[scale=0.18]{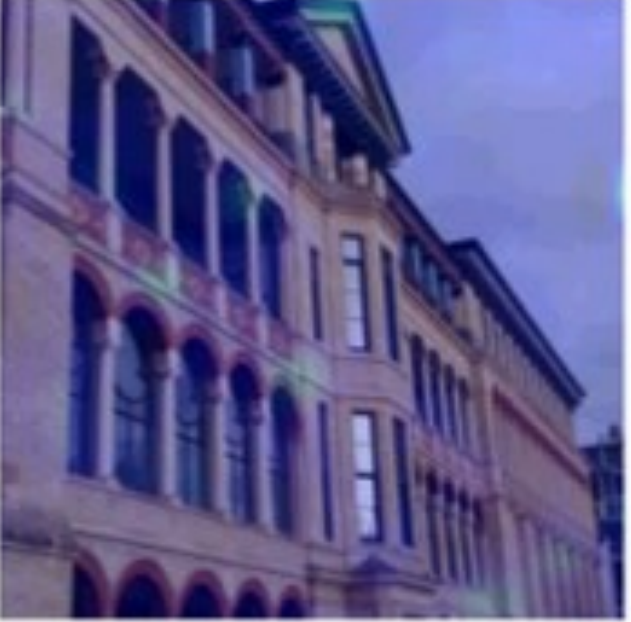}}
\subfigure[Old
Hospital]{\includegraphics[scale=0.18]{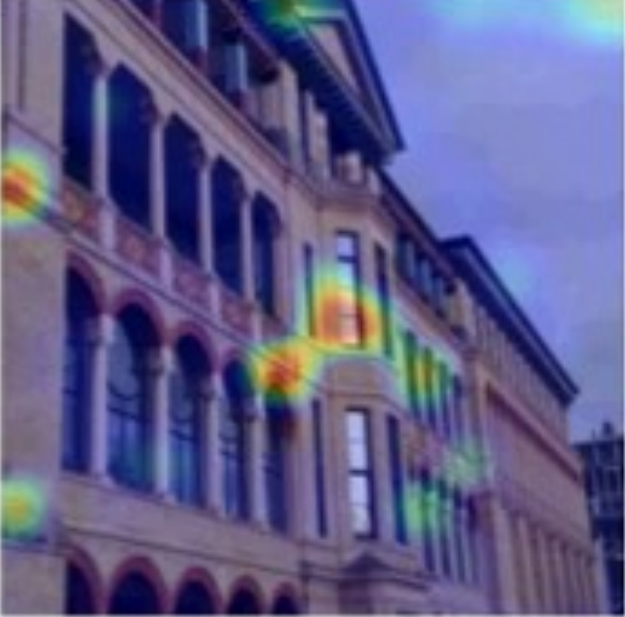}}
\subfigure[Shop
Facade]{\includegraphics[scale=0.18]{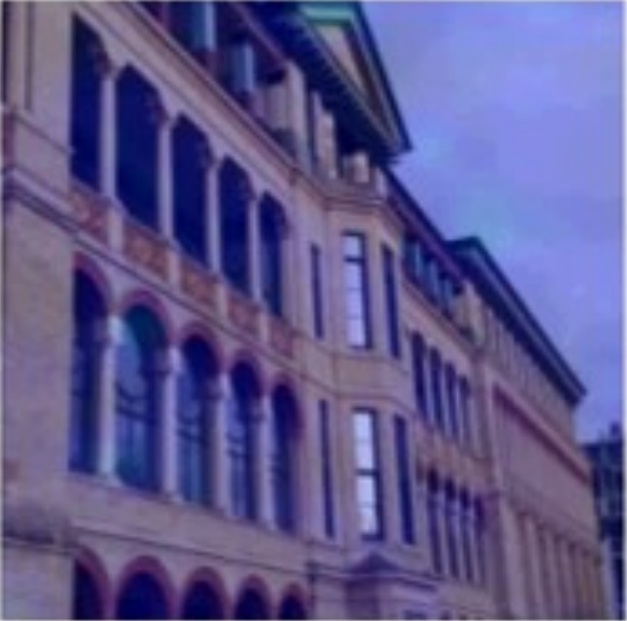}}
\subfigure[St.
Mary]{\includegraphics[scale=0.18]{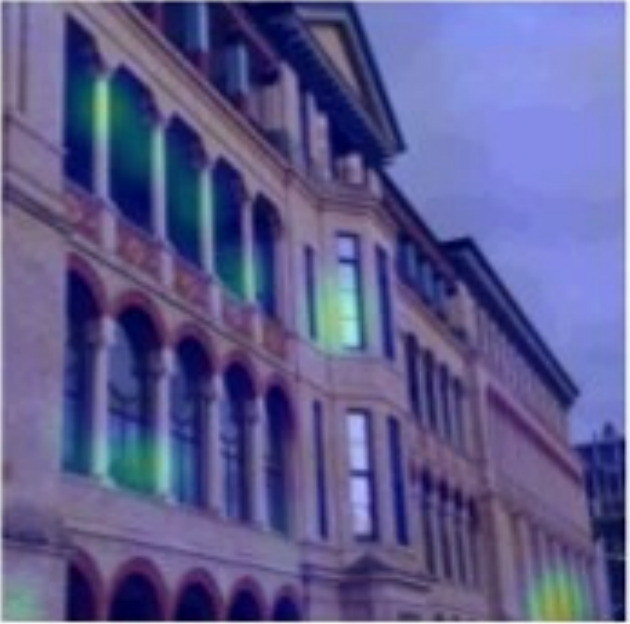}}
\end{center}
\caption{Translational Decoder attention visualization $\left\{ \mathbf{X}%
_{i}\right\} _{1}^{N}$. Each activation relates to a different scene. The
activations are due to an input image from the old Hospital scene. The
activations of the corresponding scene are notably stronger.}
\label{fig:att-decoder}
\end{figure}

The visualization of attention maps in attention-based schemes provides an
intuitive interpretation of the visual cues captured by the Transformer
Encoder. For this purpose, we visualize the upsampled attention weights of
the last encoder layer as heatmaps that are overlaid on the input images.
Figure \ref{fig:att-enocder} shows the attention map of an image taken from
the \textit{Chess} scene in the 7Scenes dataset. We show the activations
when training on three and on seven scenes (top and bottom row, respectively). Training on more scenes allows the
network to better capture informative image cues for both positional and
orientational embedding. In particular, positional attention focuses on
corner-like objects, while orientational attention emphasizes elongated
edges. We also visualize the attentions $\left\{ \mathbf{X}_{i}\right\}
_{1}^{N}$ at the outputs of the positional decoder for an image from the
\textit{OldHospital} scene (Fig. \ref{fig:att-decoder}). Each each
activation corresponds to a particular scene. Indeed, the activations corresponding
to the OldHospital scene (Fig. \ref{fig:att-decoder}b) are significantly
stronger. We include additional analysis of the decoder attention in the supplementary material. 

\begin{table}[tbh]
\caption{Ablations of the convolutional backbone of our model, evaluated on
the 7Scenes dataset. We report the average of median position and
orientation errors, across all scenes. The model choice is highlighted in bold.}
\label{table:ablation-cnn}\centering%
\begin{tabular}{lll}
\hline
\textbf{Backbone} & \textbf{Position } & \textbf{Orientation} \\
\multicolumn{1}{c}{} & \multicolumn{1}{c}{\textbf{[meters]}} & \textbf{%
[degrees]} \\ \hline
\multicolumn{1}{c}{Resnet50} & \multicolumn{1}{c}{0.19} & \multicolumn{1}{c}{
8.6} \\
\multicolumn{1}{c}{\textbf{EfficientNetB0}} & \multicolumn{1}{c}{0.18} &
\multicolumn{1}{c}{7.28} \\
\multicolumn{1}{c}{EfficientNetB1} & \multicolumn{1}{c}{0.17} &
\multicolumn{1}{c}{7.26} \\ \hline
\end{tabular}%
\end{table}
\begin{table}[tbh]
\caption{Ablations of activation maps evaluated on the 7Scenes dataset. $%
\mathbf{A_{x}}$ and $\mathbf{A_{q}}$ are sampled at different resolutions
and passed to the respective transformer head. We report the average of
median position and orientation errors, across all scenes. The model choice
is highlighted in bold.}
\label{tb:ablsation_map_res}\centering%
\begin{tabular}
[c]{ccc}\hline%
\begin{tabular}
[c]{@{}c}%
\textbf{Resolution}\\
\textbf{$\mathbf{A_{x}}/\mathbf{A_{q}}$}%
\end{tabular}
&
\begin{tabular}
[c]{@{}c}%
\textbf{Position}\\
\textbf{[meters]}%
\end{tabular}
&
\begin{tabular}
[c]{@{}c}%
\textbf{Orientation}\\
\textbf{[degrees]}%
\end{tabular}
\\\hline
$28$x$28$x$40$/$14$x$14$x$112$ & 0.22 & 7.47\\
$\mathbf{14}$x$\mathbf{14}$x$\mathbf{112}$\textbf{/}$\mathbf{28}$x$\mathbf{28}$x$\mathbf{40}$ &
0.18 & 7.28\\
$14$x$14$x$112$/$14$x$14$x$112$ & 0.19 & 7.78\\\hline
\end{tabular}
\end{table}
\begin{table}[tbh]
\caption{Ablations of the number of layers in the Encoder and Decoder
components, evaluated on the 7Scenes dataset. We report the average of
median position and orientation errors, across all scenes. The model choice
is highlighted in bold.}
\label{table:ablation-encoder-decoder-layers}\centering%
\begin{tabular}{lll}
\hline
\textbf{Encoder/Decoder} & \textbf{Position } & \textbf{Orientation} \\
\multicolumn{1}{c}{\textbf{\# Layers}} & \multicolumn{1}{c}{\textbf{[meters]}
} & \textbf{[degrees]} \\ \hline
\multicolumn{1}{c}{2} & \multicolumn{1}{c}{0.19} & \multicolumn{1}{c}{7.48}
\\
\multicolumn{1}{c}{4} & \multicolumn{1}{c}{0.18} & \multicolumn{1}{c}{6.94}
\\
\multicolumn{1}{c}{\textbf{6}} & \multicolumn{1}{c}{0.18} &
\multicolumn{1}{c}{7.28} \\
\multicolumn{1}{c}{8} & \multicolumn{1}{c}{0.18} & \multicolumn{1}{c}{6.92}
\\ \hline
\end{tabular}%
\end{table}
\begin{table}[tbh]
\caption{Ablations of the transformer's latent dimension, $C_{d}$, evaluated
on the 7Scenes dataset. We report the average of median position and
orientation errors, across all scenes. The model choice is highlighted in
bold.}
\label{table:ablation-transformer-dims}\centering
\begin{tabular}{lll}
\hline
\textbf{Transformer Dimension} & \textbf{Position } & \textbf{Orientation}
\\
\multicolumn{1}{c}{} & \multicolumn{1}{c}{\textbf{[meters]}} & \textbf{%
[degrees]} \\ \hline
\multicolumn{1}{c}{64} & \multicolumn{1}{c}{0.18} & \multicolumn{1}{c}{8.06}
\\
\multicolumn{1}{c}{128} & \multicolumn{1}{c}{0.19} & \multicolumn{1}{c}{7.56}
\\
\multicolumn{1}{c}{\textbf{256}} & \multicolumn{1}{c}{0.18} &
\multicolumn{1}{c}{7.28} \\
\multicolumn{1}{c}{512} & \multicolumn{1}{c}{0.18} & \multicolumn{1}{c}{7.19}
\\ \hline
\end{tabular}%
\end{table}

\subsection{Ablation Study}

 In order to study the effect of different architecture design choices, we
conduct multiple ablation experiments on the 7Scenes dataset (Tables \ref%
{table:ablation-cnn}-\ref{table:ablation-transformer-dims}). On each
experiment, we start from the architecture used for our comparative analysis
(Section \ref{sec:comparative_analysis}) and modify a single algorithmic
component/hyper-parameter. We compute the median position and orientation
errors for each scene and report the average across scenes. Our ablation
study focuses on two main aspects of our approach: (a) the derivation of
activation maps (backbone and resolution) and (b) the transformer
architecture.\newline
\textbf{Convolutional Backbone.} We consider three convolutional encoders
for our backbone choice: ResNet50, EfficientNetB0 and EfficientNetB1. The
results obtained with these backbones are shown in Table \ref%
{table:ablation-cnn}. The two EfficientNet variants achieve a better
performance compared to the ResNet50 backbone, either due to overfitting
(26M parameters for ResNet50 compared to 5.3M and 7.8M for EfficientNetB0
and EfficientNetB1, respectively \cite{tan2019efficientnet}) or a better
learning capacity \cite{tan2019efficientnet}. The best performance is
achieved with the EfficientNetB1 backbone, suggesting that further
improvements in accuracy can be obtained with appropriate deeper models
(e.g., deeper EfficientNet models), on the expense of memory and runtime.%
\newline
\textbf{Resolution of Activation Maps.} The EfficientNet backbone can be
sampled at different endpoints. As we move along these endpoints, the
receptive field and the depth of each entry grow. Thus, activation maps
sampled at different levels capture different features which may vary in how
informative they are for position and orientation estimation. In order to
evaluate this effect, we train our model by sampling the position and
orientation activation maps, $A_{x}$ and $A_{q}$, at different endpoints.
Specifically, we consider sampling both $A_{x}$ and $A_{q}$ from the same
endpoint, with a resolution of $14\times14\times112$ or when segregating the
sampling from two different resolutions: $14\times14\times112$ and $%
28\times28\times40$. Table \ref{tb:ablsation_map_res} shows the results of
these three combinations. The best performance is obtained when providing a
combination of coarse and fine activation maps, for the position and
orientation transformers, respectively.\newline
\textbf{Transformer Architecture} 
The main hyper-parameters of our Transformer architecture follow the standard choice.
Thus, we further evaluate the sensitivity of our
model performance to changes in 
two main hyper-parameters: the number of layers in the encoder and decoder
components and the transformer dimension, $C_{d}$. The results are shown in
Table \ref{table:ablation-encoder-decoder-layers} and Table \ref%
{table:ablation-transformer-dims}, respectively. All the considered
variants, in terms of layer number and transformer dimension, maintain SOTA
position and orientation accuracy, compared to other APR solutions (Table %
\ref{tb:7scenes_rank}). The performance improves with the Transformer
dimension, suggesting that larger models may achieve further improvement to
localization accuracy. Models with four and eight layers perform better than
the standard six layers choice for the decoder and encoder, but all variants
achieve a better performance compared to other solutions (Table \ref%
{tb:7scenes_rank}).

\section{Conclusions}

In this work we proposed a novel transformer-based approach for multi-scene
absolute pose regression. Self-attention using two Transformer Encoders
attends separately to the positional and orientational informative image
cues. Thus, aggregating the activation maps computed by the backbone CNN in
a task-adaptive way. Our formulation allows to agglomerate
non-scene-specific information in the backbone CNN and Transformer Encoders.
The scene-specific information is encoded by Transformer-Decoders, and is
queried per scene. Our approach is shown to provide a new state-of-the-art
localization accuracy for both single and multi-scene absolute regression
approaches, across outdoor and indoor datasets.

{\small
\bibliographystyle{ieee_fullname}
\bibliography{MSTransformer_arxiv}
}

\end{document}